\documentclass{article}

\usepackage[preprint]{neurips_2023}

\usepackage{natbib}
\setcitestyle{numbers,square}

\usepackage[utf8]{inputenc} 
\usepackage[T1]{fontenc}    
\usepackage{hyperref}       
\usepackage{url}            
\usepackage{booktabs}       
\usepackage{amsfonts}       
\usepackage{nicefrac}       
\usepackage{microtype}      
\usepackage{xcolor}         

\usepackage{times}
\usepackage{epsfig}
\usepackage{graphicx}
\usepackage{amsmath}
\usepackage{amssymb}

\usepackage{amsmath}
\usepackage{amssymb}
\usepackage{mathtools}
\usepackage{amsthm}

\usepackage{multirow}
\newtheorem{remark}{Remark}
\newtheorem{proposition}{Proposition}%

\usepackage{algorithm}
\usepackage[]{algpseudocode}

\title{ Improved Belief-Attention in Vision Tasks }

\author{%
  Guoqiang Zhang  \\
  University of Exeter \\
  \texttt{guoqiang.x.zhang@gmail.com} \\
}

\begin{document}

\maketitle

\begin{abstract}
Recently, Belief-Attention \cite{Guoqiang25BeliefAttention} has been proposed by first performing an orthogonal projection of the softmax-based weighted summation of $V$ vectors with respect to the original $V$ vectors and then taking the perpendicular component as the residual signal in Transformer for performance improvement. In this paper, we first conduct an ablation study showing the projected component also carries information about the token correlation, which should not be ignored. We then propose to extend Belief-Attention by making use of both the perpendicular and  projected components. In particular, the projected component goes through certain activation function and then a linear mapping before merging with the considered token. Conceptually speaking, the neural block for the projected component can be viewed as a two-layer feedforward network (FFN) within the new attention block. It is also noted that standard attention captures the token correlation via the inner-product matrix $QK^T$. We propose to introduce an additional inner-product matrix $ZZ^T$ to $QK^T$ to capture richer token correlation.  We refer to the new module as Belief2-Attention. It can be easily shown that Belief2-Attention is more expressive than standard Attention. We then verify the effectiveness of Belief2-Attention for vision tasks of image classification and segmentation.

\end{abstract}

\section{Introduction}
\label{sec:intro}


In the last decade, Transformers \citep{Transformer17} have driven major progress across diverse areas of data analysis, including natural language processing (NLP) \citep{Achiam23GPT34,Touvron23Llama2}, computer vision \citep{Dosovitskiy21ViT}, image generation and editing \citep{Peebles23ViTDiffusion, Hatamizadeh24DiffiT,Aida23BDIA}, and audio processing \citep{Latif23audioTrans}. At the core of these models lies the attention mechanism, which allows them to capture long-range dependencies within sequences. This is achieved by forming a weighted summation of value ($V$) vectors, where the weights are derived by computing the inner-product between query ($Q$) and key ($K$) vectors, followed by the softmax operation. In essence, attention enables each token to selectively incorporate information from all other tokens. After the attention step, a feedforward network (FFN) operates on each token independently, serving as a form of localized information processing. More recently, large language models (LLMs) exploit a so-called mixture of experts (MoE) as an extension of basic FFN to improve the performance, where at the inference stage, only certain percentage of weights in the FFN layer are activated depending on the particular input.

One prominent research direction focuses on reducing the quadratic computational complexity inherent in the standard attention layer when processing long token sequences. Various simplified attention schemes have been proposed, which include, for example, LinFormer \citep{Wang20LinFormer}, LongFormer \citep{Beltagy20LongFormer}, ReFormer \citep{Kitaev20Reformer},  FlashAttention \citep{Dao23Falshattion}, RingAttention \citep{Liu23RingAtten}, 
BurstAttention \citep{Sun24BurstAttention}.  FlashAttention is being widely used in practical situations as it reduces the computational complexity considerably without introducing any approximation in the standard attention layer.


\begin{figure}
\centering
\begin{minipage}{.4\textwidth}
  \centering
\includegraphics[width=0.9\linewidth]{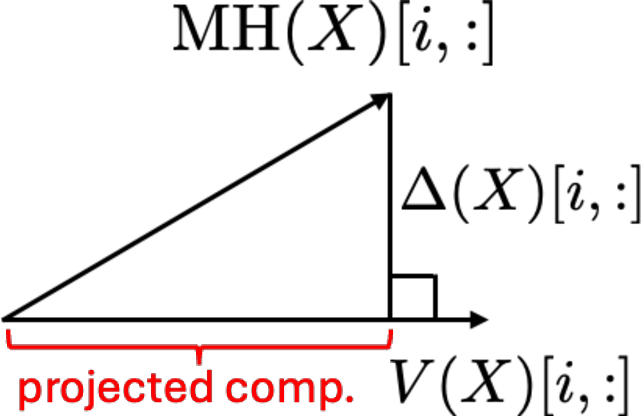}
  \caption{\footnotesize Orthogonal projection. }
  \label{fig:orthon}
\end{minipage}%
\begin{minipage}{.60\textwidth}
  \centering
\includegraphics[width=1.0\linewidth]{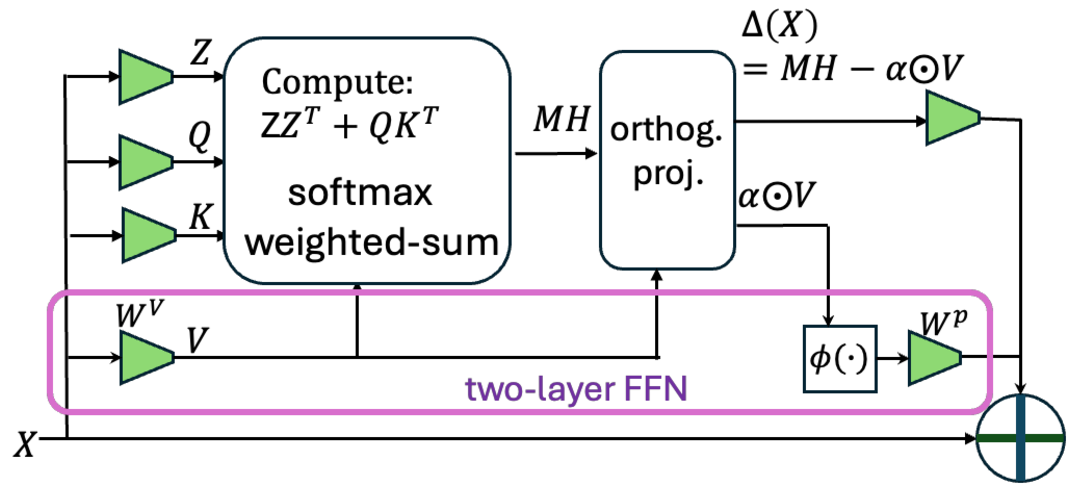}
  \caption{\footnotesize Demonstration of Belief2-Attention, where $MH$ refers to ``multi-head attention".  The two-layer FFN consists of two linear mappings, one for computing $V$ from $X$ and the other one for post-processing $\phi(\alpha\odot V)$ where $\phi(\cdot)$ is an (element-wise) activation function. }
  \label{fig:belief2_demo}
\end{minipage}
\vspace{-5mm}
\end{figure}

Another important research trend attempts to modify the attention layer in Transformer for better performance rather than for reduction of computational complexity.  For instance, the work \cite{Guoqiang25BeliefAttention} proposes a so-called Belief-Attention (see Section~\ref{sec:pre} for details). The basic idea is to first performing an orthogonal projection of the softmax-based weighted summation of $V$ vectors onto the original $V$ vectors and then taking the perpendicular component as the residual signal. On the contrary, the standard attention takes the softmax-based weighted summation of $V$ vectors directly as the residual signal. It is argued in \cite{Guoqiang25BeliefAttention} that the perpendicular component after orthogonal projection reflects the discrepancy between the weighted summation of $V$ vectors and the $V$ vectors themselves, making the tokens update more along the tangent directions and less according to their magnitudes. The recent work \cite{Zhai26ESA} also recommends to take the perpendicular component as the residual signal in an attention layer. The main difference between \cite{Guoqiang25BeliefAttention} and \cite{Zhai26ESA} is that the former considers both per-attention-head and global perpendicular components, while the latter only considers the per-attention-head perpendicular component (see Remark~\ref{remark:ESA_beliefattention} for a detailed discussion).

In this work, we first perform an ablation study of Belief-Attention, arguing that the projected component after orthogonal projection also carries information of token correlation,  which should not be abandoned. Based on the above observation, we propose letting the projected component first goes through certain activation (can be an identity function if needed) function and then a linear mapping before merging with the considered token. From a high-level viewpoint, the newly proposed attention layer includes a two-layer FFN for processing the projected component (see Fig.~\ref{fig:belief2_demo} for demonstration). We further propose to introduce an additional term to the inner product of $Q$ and $K$ vectors, which is computed by taking the inner product of $Z$ vectors and themselves for capturing richer token correlation. We refer to the new attention layer as \emph{Belief2-Attention}.  In brief, our main contributions can be summarized to be: 
\begin{enumerate}
    \item Performing an ablation study of Belief-Attention \cite{Guoqiang25BeliefAttention} showing that the projected component (see Fig.~\ref{fig:orthon} and \ref{fig:residual_impact}) should not abandoned. 
    \item Proposing Belief2-Attention by first introducing a two-layer FFN regarding the projected component and then computing and including the inner-product of $Z$ vectors and themselves to the QK-based term before the softmax operation (see Fig.~\ref{fig:belief2_demo}). We note that the second step of including the inner-product of $Z$ vectors and themselves is optional and should depend on particular applications (see Subsection~\ref{sub:Z_innerproduct}). 
    \item Experiments on image classification and segmentation show that Belief2-Attention brings considerable performance gain over standard Attention. On the other hand, Belief-Attention does not always produce performance gain over standard Attention (e.g., image segmentation for image classification). We notice that Belief2-Attention needs to introduce additional parameters. To keep the total number of model parameters remain the same whenever possible,\footnote{\small In the experiment for image segmentation of the original \href{https://github.com/rstrudel/segmenter}{\underline{open-source}}, a pretrained model is first loaded before training. It is found that if the parameters of the FFN layers are manually reduced by throwing away some sub-weight-matrices, the performance of Belief2-Attention is dropped significantly. We therefore do not reduce the parameters of the FFN layers. }  the number of parameters in each FFN layer is reduced accordingly. 
\end{enumerate}

\section{Ablation Study of Belief-Attention } 
\label{sec:pre}

In this section, we first briefly revisit the Belief-Attention in \cite{Guoqiang25BeliefAttention}. We then perform an ablation study of Belief-Attention for image classification over CIFAR100. We will show by experiment that the projected component after orthogonal projection still carries information of token correlation, which should not be ignored by Belief-Attention.  

\subsection{Revisiting Belief-Attention in \cite{Guoqiang25BeliefAttention} }
\label{sub:beliefattention}
\vspace{-0mm}

As mentioned earlier, Belief-Attention includes both the per-attention-head and global perpendicular components as the residual signals before being processed by a linear mapping. Following the convention of python-based implementation (e.g., pytorch) of attention, suppose a tensor $X\in \mathbb{R}^{n\times d}$ of $n$ tokens is the input from the layer below in a Transformer, where each token is of dimension $d$.  We use the row vector $X[i,:]$ to denote the $i$th token. The update expression of Belief-Attention  can be represented as (see  \citep{Guoqiang25BeliefAttention, MHA23Pytorch, Dosovitskiy21ViT}) 
\begin{align}
\textrm{H}_m(X) &\hspace{-0.5mm}=\hspace{-0.5mm} \textrm{attention}(\overbrace{XW_m^{Q}}^{Q_m}, \overbrace{XW_m^{K}}^{K_m}, \overbrace{XW_m^V}^{V_m}) \label{equ:att_ffn1} \\
\Delta_m(X)[i,:] &= \underbrace{H_m(X)[i,:] - \overbrace{\beta_{m,i}V_m[i,:]}^{\textcolor{blue}{\textrm{projected comp.}}}}_{\textcolor{blue}{\textrm{perpendicular comp.}}} \quad \beta_{m,i} = \frac{\langle  \textrm{H}_m(X)[i,:], V_m[i,:]\rangle}{\langle V_m[i,:],V_m[i,:] \rangle }
\label{equ:sub_orth} \\
\textrm{MH}(X) &= \textrm{Concat}(\textrm{H}_1(X),\ldots, \textrm{H}_M(X)) \label{equ:att_ffn2} \\
V(X) &=\textrm{Concat}(V_1,\ldots, V_M)  \label{equ:att_ffn_V}
\\
{\Delta}^l(X)&=\textrm{Concat}(\Delta_1(X), \ldots, \Delta_M(X)) \label{equ:att_ffn_delta}
\\
\Delta(X)[i,:]&=\underbrace{\textrm{MH}(X)[i,:] - \overbrace{\alpha_i V(X)[i,:]}^{\textcolor{blue}{\textrm{projected comp.}}}}_{\textcolor{blue}{\textrm{perpendicular comp.}}} \quad \alpha_i = \frac{\langle  \textrm{MH}(X)[i,:], V(X)[i,:]\rangle}{\langle V(X)[i,:],V(X)[i,:]} \rangle \label{equ:sub_orth_global} \\
X &\Leftarrow {X+{\Delta}(X)W+{\Delta}^l(X)W^l},\label{equ:belief_attn_plus} 
\end{align}
where pre-normalization is omitted for simplicity, $i\in \{1,2,\ldots, n\}$ denotes the $i$th token, $m\in \{1,2\ldots, M\}$ denotes the $m$th attention-head, $(W_m^Q, W_m^K, W_m^V)$ are the three learnable matrices for computing $(Q_m, K_m, V_m)\in (\mathbb{R}^{n\times d_m},\mathbb{R}^{n\times  d_m},\mathbb{R}^{n\times d_m})$ of the $m$th attention. 
The notations $\textrm{H}$ and $\textrm{MH}$ stand for ``head" and ``multi-head", respectively.  

By inspection of (\ref{equ:att_ffn1})-(\ref{equ:belief_attn_plus}), we can conclude that $\Delta_m(X)$ is the local perpendicular component for the $m$th attention head and $\Delta$ is the global perpendicular component after projecting each $MH(X)[i,:]$ onto its associated $V(X)[i,:]$ vector in a token-wise manner. To facilitate discussion later on, we rewrite (\ref{equ:sub_orth}) and (\ref{equ:sub_orth_global}) in a compact form as: 
\begin{align}
\Delta_m(X) =& H_m(X) - \overbrace{\beta_m \odot V_m}^{\textcolor{blue}{\textrm{projected comp.}}}   \quad m=1,\ldots, M \label{equ:local_per_compact} \\
\Delta(X) =& H_m(X) - \underbrace{\alpha \odot V}_{\textcolor{blue}{\textrm{projected comp.}}},
 \label{equ:global_per_compact}
\end{align}
where $\alpha = [\alpha_1,\ldots, \alpha_n]$, $\beta_m=[\beta_{m,1}, \ldots, \beta_{m,n}]$, and $\odot$ denotes elementwise multiplication. 
Finally, the tokens in $X$ are updated by making use of both the local $\Delta^l$ and global perpendicular components $\Delta$ as given by (\ref{equ:belief_attn_plus}).

We now briefly study the attention operation in Equ.~(\ref{equ:att_ffn1}). It is well-known that $H_m$ is a QK-softmax-based weighted summation of the $n$ row vectors in $V_m$, given by 
\begin{align}
\textrm{H}_m(X) = \overbrace{\textrm{softmax}\left(\frac{Q_mK_m^T}{\sqrt{d_m}}\right)V_m}^{\textrm{\textcolor{blue}{info. aggregation}}}
\label{equ:att_ffn2_6}
\end{align}
where $d_m$ is the dimension of the row vectors in $Q_m$.  The softmax term computes the unified relevance of each token $i$ with respect to all other tokens after obtaining the inner-products of the current query  vector $Q_m[i,:]$ with all other key vectors in $\{K_m[j,:]\}_{j=1}^n$, which generally stabilizes the training process in comparison to other forms of weighted summation. 

Finally, we  notice that Belief-Attention needs to introduce an additional weight matrix $W^l$ in (\ref{equ:belief_attn_plus}) in comparison to the standard attention. In general, the matrix $W^l$ per attention layer would incur only a small overhead in model parameters.. 

\begin{remark}
\label{remark:ESA_beliefattention}
It is worth noting that the recent work \cite{Zhai26ESA} considers taking the local perpendicular component $\Delta^l(X)$ in (\ref{equ:att_ffn_delta}) as the residual signal in an attention layer and observe noticeable performance gain in training various sizes of GPT2 for natural language processing (NLP). As a result, the work of \cite{Zhai26ESA} can be considered as a special case of Belief-Attention by setting $W=0$ in (\ref{equ:belief_attn_plus}).
\end{remark}

\subsection{Towards understanding impact of projected component in a Transformer}

We note from Subsection~\ref{sub:beliefattention} that Belief-Attention only takes the two perpendicular components $\Delta(X)$ and $\Delta^l(X)$ as the residual signals when updating the tokens as indicated in (\ref{equ:belief_attn_plus}). One natural question is if the local $\{\beta_m\odot V_m\}_{m=1}^M$ and global projected components $\alpha\odot V$ in (\ref{equ:local_per_compact})-(\ref{equ:global_per_compact}) also carry useful information about the token correlations or not. In other words, if the coefficients $\{\beta_m\}_{m=1}^M$ and $\alpha$ reflect the token correlation to a noticeable level or not.  

In order to gain insights regarding the above question, we conducted an ablation study evaluating the impact of different residual signals in the attention layers of a Transformer on the validation accuracy when training the models over CIFAR100 for image classification. 

\begin{figure}
\centering
\vspace{0mm}\includegraphics[width=100mm]{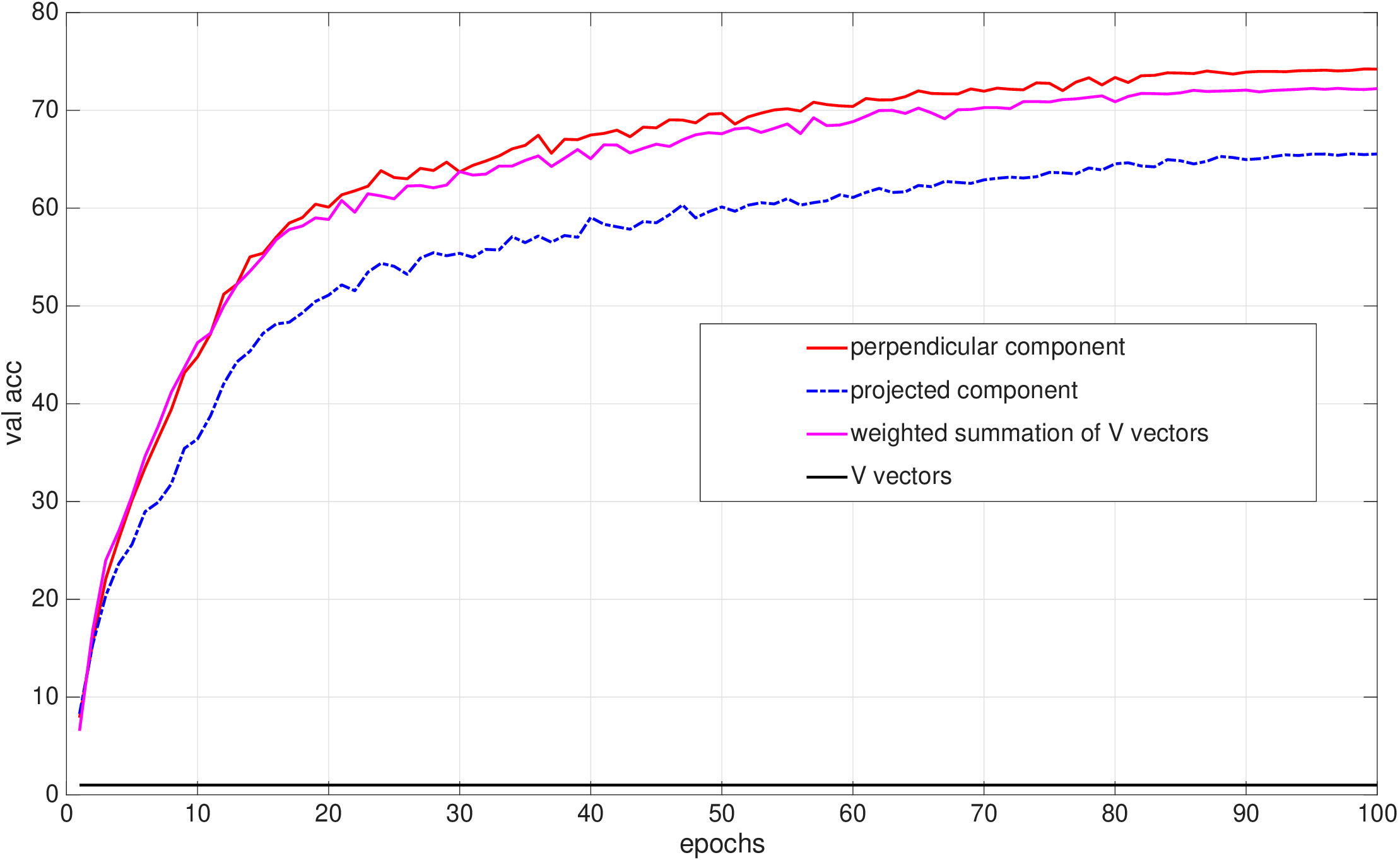}
\vspace{-3mm}
\caption{\small Impact of different residual signals in the attention layers of a Transformer being trained over CIFAR100 for image classification.  The curve for weighted summation of $V$ vectors is in fact the performance of the standard Attention layer. The first open-source in Table~\ref{tab:opensources} is utilized for this experiment.}
\label{fig:residual_impact}
\end{figure}

Fig.~\ref{fig:residual_impact} visualizes the validation accuracy over epochs for different residual signals being employed in all the attention layers of a Transformer. It is clear from the figure that both the perpendicular and projected components lead to reasonable validation performance. The results also suggest that the perpendicular component is more informative of the token-correlation than the projected components. 

Fig.~\ref{fig:residual_impact} also includes the validation performance (the curve in black color) by taking the original $V$ vectors as the residual signals in all the attention layers of the Transformer. Unsurprisingly, the model fails to obtain any meaningful classification results regardless of the number of training epochs.     

In summary, the above results indicate that the coefficients $\{\beta_m\}_{m=1}^M$ and $\alpha$ in the projected components indeed reflect the token correlation to a noticeable level. The next question is how to modify Belief-Attention properly in order to make use of both the perpendicular and projected components effectively for performance improvement. 

\section{Belief2-Attention}
\label{sec:propose}

In this section, we first motivate the benefits of introducing a two-layer FFN within Belief2-Attention for handling the projected components. We then briefly explain that the softmax operation could also be properly modified for performance improvement by introducing the inner-product of $Z$ vectors and themselves if the machine learning tasks considered allow this. After that, we present and analyze  the update expressions of Belief2-Attention. We will argue that under certain mild conditions, Belief2-Attention has higher representational capability than the standard Attention.

\subsection{Motivation of two-layer FFN within Belief2-Attention (Compulsory)} 

As demonstrated in Fig.~\ref{fig:belief2_demo}, we propose to let the projected component $\alpha \odot V$ first go through an (element-wise) activation function $\phi(\cdot)$ and then a linear mapping $W^p$ before merging with the considered tokens. From a high level viewpoint, a two-layer FFN is implicitly embedded within Belief2-Attention. In particular, the FFN has two linear mapping, of which the first mapping $W^V=\textrm{Concat}(W^V_1,\ldots, W^V_M)$ converts the tokens in $X$ into value vectors in $V$ (see (\ref{equ:att_ffn1})) and the second mapping $W^p$ post-process $\phi(\alpha\odot V)$ before merging with the considered tokens.  

In principle, with the two-layer FFN, Belief2-Attention would have greater ability to modify tokens as desired than both Belief-Attention and the standard Attention. As a result, when building a Transformer by stacking a set of Belief2-Attention and traditional FFN layers, the final model would have high capability to fit the input-output training data.  


\subsection{Motivation of $ZZ^T+QK^T$ before softmax operation (Optional) } 
\label{sub:Z_innerproduct}

We notice from (\ref{equ:att_ffn2_6}) that the standard Attention first computes $QK^T$ and then apply the softmax operation in a row-wise manner to obtain the weighting scores. Since $Q$ and $K$ matrices are different, the resulting inner-product matrix $QK^T$ is not symmetric. As a result, after applying the (row-wise) softmax operation, the relative weighting score of token $i$ to token $j$ would be different from that of token $j$ to token $i$. 

We propose to introduce an additional inner-product matrix $ZZ^T$ into $QK^T$ to capture richer token correlations. The additional term $ZZ^T$ would naturally make the resulting weighting scores a bit more symmetric. Since the weight matrix $W^Z$ behind the tensor $Z$ is learnable, the relative weighting scores will be learned dynamically from the input-output data. Theoretically, in the worst case, $W^Z$ can learn to approach 0 to remove the impact of $ZZ^T$. 

In practice, not all machine learning tasks may benefit from imposing $ZZ^T$ onto the $QK^T$ inner-product matrix. For instance, one common scenario is to load a pretrained Transformer model and further train it on a downstream task. In this case, the weight matrices $W^Q$ and $W^K$ for $Q$ and $K$ tensors are already well trained from before. On the other hand, the weight matrix for the $Z$ tensor need to be trained from scratch for the downstream task. This might not bring performance gain due to the imbalanced training setup.

\subsection{Update expressions}
\label{sub:lookahead}

Based on the earlier motivations for Belief2-Attention,  its update expressions are designed to be
\begin{align}
\textrm{H}_m(X) &\hspace{-0.5mm}=\hspace{-0.5mm} \textrm{attention}(\overbrace{XW_m^{Z}}^{Z_m}, \overbrace{XW_m^{Q}}^{Q_m}, \overbrace{XW_m^{K}}^{K_m}, \overbrace{XW_m^V}^{V_m}) = \textrm{softmax}\left(\frac{Q_mK_m^T+ \textcolor{blue}{Z_mZ_m^{T}} }{\sqrt{d_m}}\right)V_m \label{equ:belief2_1} \\
\textrm{MH}(X) &= \textrm{Concat}(\textrm{H}_1(X),\ldots, \textrm{H}_M(X)) \label{equ:belief2_2} \\
V(X) &=\textrm{Concat}(V_1,\ldots, V_M)  \label{equ:belief2_3}
\end{align}
\begin{align}
\hspace{0mm}\Delta(X)&=\textrm{MH}(X) - \alpha \odot V(X) \label{equ:belief2_global} \\
\hspace{0mm}X &\Leftarrow {X+{\Delta}(X)W+\textcolor{blue}{{\phi}(\alpha \odot V(X))} W^p }, \textcolor{white}{\textrm{ asdf asdfsa fsa saf as   asfd asfs asfda a saf safs}} \label{equ:belief2_5}  
\end{align}
where we use $\phi(\cdot)$ to denote an (linear/nonlinear) activation function, each element $\alpha_i$ in $\alpha$ is computed by following (\ref{equ:sub_orth_global}). As we discussed earlier, one can optionally introduce $ZZ^{T}$ into the original inner-product matrix $QK^T$ depending on the particular applications.

\textbf{On relationship between Belief2-Attention and standard Attention}:  We now argue that Belief2-Attention has greater representational capacity than the standard Attention under certain mild conditions. Firstly, we note that (\ref{equ:belief2_5}) can be rewritten as 
\begin{align}
X &\Leftarrow {X+  \overbrace{(MH(X) - \alpha \odot V(X))}^{\textcolor{blue}{\textrm{perpendicular comp.}}} W +{\phi}(\overbrace{\alpha \odot V(X))}^{\textcolor{blue}{\textrm{projected comp.}}}} W^p,\label{equ:belief2_6} \\
&= X+  (MH(X) - \alpha \odot V(X)) W + \underbrace{[{\phi}(\alpha \odot X W^V)] W^p}_{\textcolor{blue}{\textrm{two-layar FFN}}},\label{equ:belief2_66}
\end{align}
where unlike that of Belief-Attention, both the perpendicular and projected components are present in the expression. No useful information is dropped on purpose in the above expression.   

If we let $\phi(\cdot)$ be an identity (no nonlinear mapping is enforced) function and $W^p=W$, it is immediate that (\ref{equ:belief2_6}) reduces to the update expression for the standard Attention:
\begin{align}
X &\Leftarrow {X+ MH(X)W} \label{equ:belief2_7}.
\end{align}
This indicates that the functional space of Belief2-Attention includes that of the standard Attention. In other words, Belief2-Attention indeed has higher representational capacity. We use $\mathcal{P}[f_{\theta}]$ to denote the functional space parameterized by $\theta$. The above results can be summarized in a proposition below:  

\begin{proposition}
Let $\phi(\cdot)$ be an identity function. Suppose the update expression for Belief2-Attention follows from (\ref{equ:belief2_1})-(\ref{equ:belief2_5}) while the update expression for the standard Attention is the one by setting $W^p=W$. We then have 
\begin{align}
\mathcal{P}[\textrm{Attention}] \subseteq \mathcal{P}[\textrm{Belief2-Attention}].
\end{align}
\end{proposition}

Next, we discuss the impact of $W$ and $W^p$ in (\ref{equ:belief2_6}) by letting $\phi(\cdot)$ be an identity function. In this case, the two linear mappings operate on the perpendicular and projected components independently.  Naturally, (\ref{equ:belief2_6}) has more degree of freedom to modify tokens as desired than that of (\ref{equ:belief2_7}) in the standard Attention. It is expected that the training loss incurred by Belief2-Attention would be lower than that of the standard Attention. As will be discussed later on,  experiments on segmentation and image classification over ImageNet indeed verify the above statement.

\begin{remark}
It is not known to us at the moment if the activation function $\phi(\cdot)$ can be dynamically and effectively learned or not in Belief2-Attention. One future research work would be to investigate how to learn the activation function $\phi(\cdot)$ in the training process smoothly instead of manual setup, which would make Belief2-Attention much more powerful.   
\end{remark}

\textbf{Study of two-layer FFN in (\ref{equ:belief2_66})}: It is worth noting that the two-layer FFN in (\ref{equ:belief2_66}) is slightly different from the conventional two-layer FFN in a Transformer, which can be generally represented as 
\begin{align}
\textrm{FFN}(X)=[{\varphi}(X W^1)] W^2, \label{equ:ffn}
\end{align}
where $\varphi$ denotes an element-wise activation function, and $W^1$ and $W^2$ are the two linear mapping functions.    

By inspection of (\ref{equ:belief2_66}) and (\ref{equ:ffn}), it is seen that the two-layer FFN in Belief2-Attention is more informative. The weighting vector $\alpha$ before $XW^V$ in (\ref{equ:belief2_66})  carries information of token correlation. On the contrary, the conventional FFN does not have such information.

\subsection{Parameter and computational overhead of Belief2-Attention} 
\label{sub:overhead}

One can easily conclude from (\ref{equ:belief2_1})-(\ref{equ:belief2_5}) that Belief2-Attention has to introduce two additional weight matrices $W^p$ and $W^Z$ in comparison to the standard Attention. Since Belief2-Attention implicitly includes a two-layer FFN, one can reduce the number of parameters in the conventional FFN layers to keep the total number of model parameters the same as the standard Transformer whenever possible.     

Belief2-Attention would also introduce a computational overhead due to the orthogonal projection and the two-layer FFN. On the other hand, the computational cost of the conventional FFN layers would be slightly reduced due to the corresponding reduction in the number of model parameters.  See Table~\ref{tab:imagenet}, \ref{tab:cifar100}, \ref{tab:ablation_cifar100}, and \ref{tab:ablation_imagenet} for information about the computational overhead.

\section{Experimental Results}
\label{sec:exp}

We evaluated Belief2-Attention for three vision tasks: (1) image segmentation  over the ADE20K dataset\footnote{\href{dataset link}{https://github.com/CSAILVision/ADE20K}}; (2) image classification over ImageNet; (3) image classification over CIFAR100.  All the experiments were conducted on a computer with a single Nvidia Geforce A6000 GPU with 48GB memory.     

For the 2nd and 3rd tasks, since we train the Transformer models from scratch, the number of parameters in the conventional FFN layers are reduced properly to account for the parameter overhead introduced in Belief2-Attention. After parameter reallocation, it is ensured that Belief2-Attention based models have the same number of parameters as the original Transformer models. 

For the 1st task of image segmentation, a pretrained model is loaded for further training in the original open-source as listed in Table~\ref{tab:opensources}. As we mentioned earlier, we didn't  reduce the number of parameters in the conventional FFN layers of the Transformer to be able to produce reasonable performance for Belief2-Attention. We also found that there is no need to introduce the inner-product matrix $ZZ^{T}$ into $QK^T$ for the considered task. As a result, the parameter overhead of Belief2-Attention comes from the $W^p$ matrix in each attention layer. 


In brief, it is found that Belief2-Attention outperforms both the standard Attention and Belief-Attention in all three tasks. 

     \begin{figure}
     \centering
\vspace{0mm}\includegraphics[width=120mm]{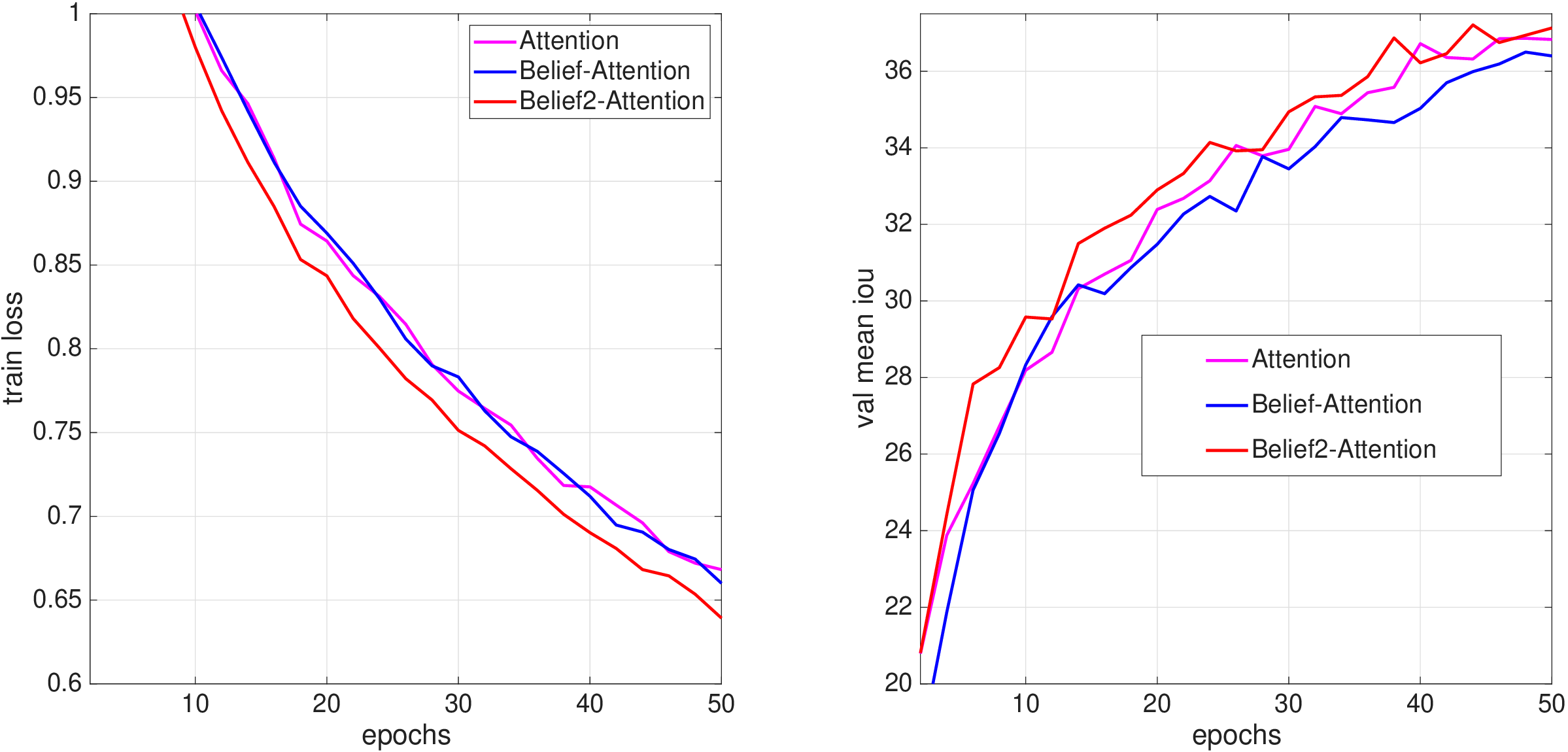}
     \caption{\footnotesize{ Performance comparison for image segmentation over the ADE20K dataset. }  }
\label{fig:seg_compare}
     \end{figure}

\subsection{Image segmentation over ADE20K}
\label{subsec:vit_seg}
The first experiment adopted the 2nd opensource in Table~\ref{tab:opensources} for evaluating Belief2-Attention, Belief-Attention, and Attention for the image segmentation task. The model type ``vit$\_$tiny$\_$patch16$\_$384" was selected and trained for performance comparison of the three different attentions.  The training setups in terms of the hyper-parameters follow directly from the original open source.

We point out that the activation function $\phi(\cdot)$ for Belief2-Attention was selected to be the identity function, implying that no element-wise nonlinear mapping is performed. The reason we chose the identity function is that a pretrained model is loaded for further training in image segmentation. If a nonlinear activation function were chosen, it may negatively affect the gradient flow.      

Fig.~\ref{fig:seg_compare} displays both the training loss and validation mean iou over epochs for performance comparison. As expected, Belief2-Attention leads to consistently lower training loss over epochs than the standard Attention. This is because the functional space of Belief2-Attention includes that of Attention, making it much easier for the model to fit the input-output training data.     

We can also conclude from the figure that Belief2-Attention produces slightly higher validation mean iou than both Belief-Attention and Attention. 

\subsection{Image classification over ImageNet}
\label{subsec:vit_imagenet}

In this experiment, we adjusted the Vision Transformer model from the opensource\footnote{https://github.com/LTH14/JiT} for the image classification purpose over ImageNet. The original open-source is for training a diffusion model for image generation. The selected model type for our experiment is JiT-B/16. We replaced each standard attention in JiT-B/16 with Belief2-Attention and Belief-Attention, respectively. The activation function $\phi(\cdot)$ for Belief2-Attention was selected to be the Sigmoid Linear Unit (SiLU) function, which is also the one being used in the conventional FFN layers of JiT. The batchsize and the learning rate were set to be 480 and 1e-4, respectively. The optimizer AdamW with the parameter setup of $(\beta_1,\beta_2, \epsilon)=(0.9, 0.999, 1e-8)$ was utilized for the training process.  All the models were trained from scratch. After training, they are evaluated via the associated validation dataset.

\begin{figure}
\centering
\vspace{0mm}\includegraphics[width=120mm]{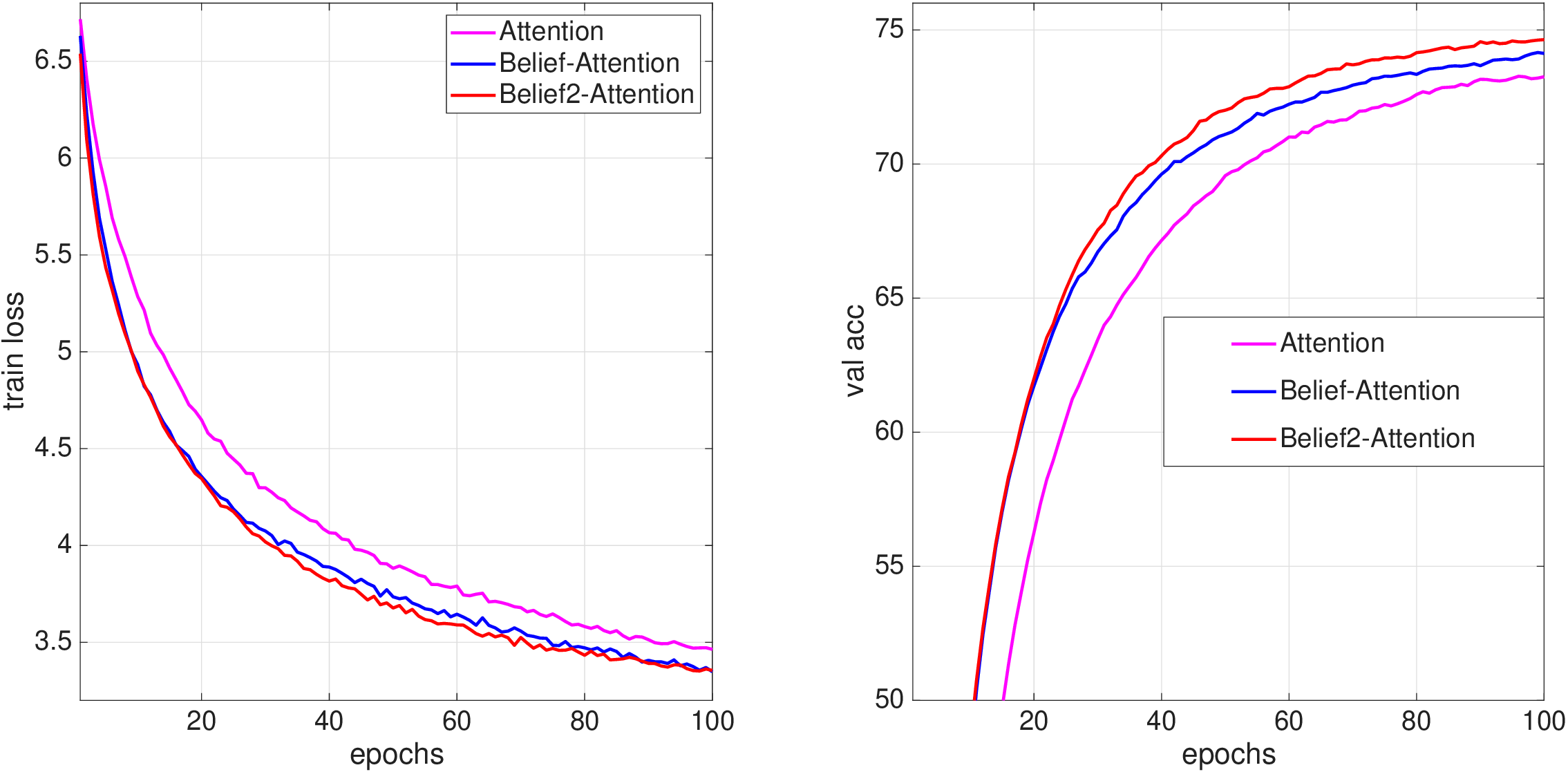}
     \caption{\footnotesize{ Performance comparison for image classification over ImageNet of 1000 classes. }  }
\label{fig:imagenet_compare}
\end{figure}

\begin{table}[h!]
\vspace*{-0.cm}
\caption{ \footnotesize{
Performance comparison of three types of attention layers for image classification over ImageNet. 
}} 
\label{tab:imagenet}
\centering
\begin{tabular}{|c||c|c|c|}
\cline{2-4} 
\multicolumn{1}{c|}{{ }}  & {\footnotesize Attention }    & {\footnotesize Belief-Attention} & {\footnotesize Belief2-Attention }    
\\ \hline
{\footnotesize val acc.} 
\hspace{-2mm}
&  
\hspace{-2mm}
{\footnotesize 73.25}
\hspace{-2mm}
& \hspace{-2mm} 
{\footnotesize 74.20} 
& {\footnotesize \textbf{74.67} } 
\\ \hline
{\footnotesize model size} 
& 
\multicolumn{3}{|c|}{\footnotesize 128.86M} 
\\ \hline
{\footnotesize time/iter (s)} 
\hspace{-2mm}
& 
\hspace{-2mm}
{\footnotesize 0.312 }
\hspace{-2mm}
& \hspace{-2mm} 
{\footnotesize 0.324 } 
& {\footnotesize 0.367} 
\\ \hline
\end{tabular}
\vspace*{-0.0cm}
\end{table}

Fig.~\ref{fig:imagenet_compare} visualizes both the training loss and validation accuracy over epochs. It is clear that both Belief2-Attention and Belief-Attention outperforms Attention  considerably as the epoch index increases. The fast convergence with Belief2-Attention based model in comparison to Attention based model can be explained by the fact that a two-layer FFN is embedded within each Belief2-Attention. This brings higher representational capability into the Belief2-Attention based model.  

Table~\ref{tab:imagenet} summarizes  information of validation accuracy, model size, and training time complexities of the three types of attention layers. It is clear that under the constraint of the same model size, Belief2-Attention achieves the best validation accuracy with slightly higher computational cost.  



\vspace{-3mm}
\subsection{Image classification over CIFAR100}
\vspace{-2mm}
In this experiment, we adopted the 1st open-source repository of Table~\ref{tab:opensources} for training ViT over CIFAR100 . We replaced the standard attention layer by Belief-Attention and Belief2-Attention for performance comparison. The number of parameters in the conventional FFN layers are properly adjusted to ensure the model size remains the same. The activation function $\phi(\cdot)$ for Belief2-Attention was set to be the Gaussian Error Linear Units (GELU) function, which is the same as the one being used in the conventional FFN layers.




Aside from the model modifications mentioned above, the training setups follow the original open-source implementation. In brief, each model was trained for 100 epoch by using the AdamW optimizer. Three experimental repetitions (with random seeds in $\{0, 50, 100\}$) were performed per training setup to mitigate the effect of randomness.

Table~\ref{tab:cifar100} summarizes the validation accuracy, model size, and training time complexities. It is clear that Belief2-Attention produces considerably higher validation accuracy than the other two attention layers at the cost of slightly higher time complexities. This indicates that the the introduced two-layer FFN in Belief2-Attention indeed helps with performance improvement.  

\begin{table}[h!]
\vspace*{-0.cm}
\caption{ \footnotesize{Performance comparison of three different attention layers for image classification over CIFAR100. 
}} 
\label{tab:cifar100}
\centering
\begin{tabular}{|c||c|c|c|}
\cline{2-4} 
\multicolumn{1}{c|}{{ }}  & {\footnotesize Attention }    & {\footnotesize Belief-Attention } & {\footnotesize Belief2-Attention }  
\\ \hline\hline
{\footnotesize val acc.} 
\hspace{-2mm}
& 
\hspace{-2mm}
{\footnotesize 72.01$\pm$0.20 }
\hspace{-2mm}
& \hspace{-2mm} 
{\footnotesize {74.18}$\pm0.25$ } 
& {\footnotesize \textbf{76.15}$\pm0.25$ } 
\\ \hline
{\footnotesize model size} 
\hspace{-2mm}
&
\multicolumn{3}{|c|}{\footnotesize 3.38M}
\\ \hline
{\footnotesize time/epoch (s)} 
\hspace{-2mm}
& 
\hspace{-2mm}
{\footnotesize 43.42 }
\hspace{-2mm}
& \hspace{-2mm} 
{\footnotesize  55.52} 
& {\footnotesize 59.74 } 
\\ \hline
\end{tabular}
\vspace*{-0.0cm}
\end{table}

\subsection{Ablation study of Belief2-Attention}

We also investigated the performance of Belief2-Attention under two special setups, which are (a): without $ZZ^T$; (b) without FFN layers. Since Belief2-Attention already includes an FFN layer by itself, it is of great interest to find out the performance of the model which is built by stacking only a set of Belief2-Attention layers.  

Table~\ref{tab:ablation_cifar100} and \ref{tab:ablation_imagenet} summarize the performance of Belief2-Attention based model under different setups. By inspection of Table~\ref{tab:imagenet}, \ref{tab:cifar100}, \ref{tab:ablation_cifar100} and \ref{tab:ablation_imagenet}, one can conclude that even without $ZZ^T$, the model performance is still better than that the standard Transformer.

\begin{table}[h!]
\vspace*{-0.cm}
\caption{ \footnotesize{
Ablation study for different setups of Belief2-Attention based ViT over CIFAR100.  
}} 
\label{tab:ablation_cifar100}
\centering
\begin{tabular}{|c||c|c|c|}
\cline{2-4} 
\multicolumn{1}{c|}{{ }}  & {\footnotesize with FFN layers and $ZZ^T$ }    & {\footnotesize without FFN layers } & {\footnotesize without $ZZ^T$  }    
\\ \hline
{\footnotesize val acc.} 
\hspace{-2mm}
&  
\hspace{-2mm}
{\footnotesize 76.15$\pm$0.25 }
\hspace{-2mm}
& \hspace{-2mm} 
{\footnotesize 72.91$\pm$ 0.85} 
& {\footnotesize {73.85}$\pm0.31$ } 
\\ \hline
{\footnotesize model size} 
\hspace{-2mm}
& 
\hspace{-2mm}
{\footnotesize 3.38M }
\hspace{-2mm}
& \hspace{-2mm} 
{\footnotesize 2.04M } 
& {\footnotesize 3.04M } 
\\ \hline
{\footnotesize time/epoch (s)} 
\hspace{-2mm}
& 
\hspace{-2mm}
{\footnotesize 59.74 }
\hspace{-2mm}
& \hspace{-2mm} 
{\footnotesize 50.23 } 
& {\footnotesize 48.96 } 
\\ \hline
\end{tabular}
\vspace*{-0.5cm}
\end{table}

\begin{table}[h!]
\vspace*{-0.cm}
\caption{ \footnotesize{
Ablation study for different setups of Belief2-Attention based ViT over ImageNet.  
}} 
\label{tab:ablation_imagenet}
\centering
\begin{tabular}{|c||c|c|}
\cline{2-3} 
\multicolumn{1}{c|}{{ }}  & {\footnotesize with FFN layers and $ZZ^T$ }  & {\footnotesize without $ZZ^T$  }    
\\ \hline
{\footnotesize val acc.} 
\hspace{-2mm}
&  
\hspace{-2mm}
{\footnotesize 74.67 }
\hspace{-2mm}
& \hspace{-2mm} 
{\footnotesize  73.73} 
\\ \hline
{\footnotesize model size} 
\hspace{-2mm}
& 
\hspace{-2mm}
{\footnotesize 128.86M }
\hspace{-2mm}
& \hspace{-2mm} 
{\footnotesize 121.78M } 
\\ \hline
{\footnotesize time/iter (s)} 
\hspace{-2mm}
& 
\hspace{-2mm}
{\footnotesize 0.367 }
\hspace{-2mm}
& \hspace{-2mm} 
{\footnotesize 0.312 } 
\\ \hline
\end{tabular}
\vspace*{-0.5cm}
\end{table}

\section{Conclusions}
In this work, we have first argued by an ablation study that the projected component after orthogonal projection in Belief-Attention carries useful information regarding token-corelation and should be not abandoned. We have then proposed Belief2-Attention by making use of both the perpendicular and projected components. A two-layer FFN is smoothly embedded in the new attention layer for handling the projected component. We also show that under certain mild conditions, Belief2-Attention has higher representational capability than the standard Attention.  Since Belief2-Attention includes a two-layer FFN already, one can reduce the number of parameters in conventional FFN layers of a Transformer. Our second contribution is to introduce an additional inner-product matrix $ZZ^T$ into the $QK^T$ matrix to capture richer token-correlation. Experimental results on image segmentation and classification show that  Belief2-Attention always outperforms the standard Attention while Belief-Attention sometimes produces inferior results.



\newpage

\appendix

\begin{table}[t!]
\begin{center}
\begin{tabular}{|l|l|}
\hline
{\footnotesize $\begin{array}{l}\textrm{CIFAR10 \&} \\ \textrm{CIFAR100}\end{array}$} & \url{https://github.com/aanna0701/SPT_LSA_ViT}
\\
\hline
{\footnotesize $\begin{array}{l}\textrm{Image segmentation  }\end{array}$} & \url{https://github.com/rstrudel/segmenter}
\\
\hline
\end{tabular}
\end{center}
\caption{\small list of open-source repositories expoited in this paper. }
\label{tab:opensources}
\end{table}

\end{document}